\documentclass{article} 
\usepackage{iclr2026_delta,times}
\usepackage{booktabs}
\usepackage[table]{xcolor}

\usepackage{amsmath,amsfonts,bm}

\def\eqref#1{equation~\ref{#1}}

\def\1{\bm{1}}

\DeclareMathAlphabet{\mathsfit}{\encodingdefault}{\sfdefault}{m}{sl}
\SetMathAlphabet{\mathsfit}{bold}{\encodingdefault}{\sfdefault}{bx}{n}

\usepackage{hyperref}
\usepackage{url}
\usepackage{graphicx} 
\usepackage{float}
\usepackage{amsthm}
\usepackage{algorithm}
\usepackage{algorithmic}
\iclrfinalcopy

\title{Discrete Meanflow Training Curriculum}

\author{Chia-Hong Hsu \\
  Department of Computer Science \\
  University of British Columbia \\
  Vancouver, BC, Canada \\
  \texttt{chsu35@student.ubc.ca} \\
  \And
  Frank Wood \\
  Department of Computer Science \\
  University of British Columbia \\
  Vancouver, BC, Canada \\
  \texttt{fwood@cs.ubc.ca}
}

\begin{document}

\maketitle

\begin{abstract}
Flow-based image generative models exhibit stable training and produce high quality samples when using multi-step sampling procedures. 
One-step generative models can produce high quality image samples but can be difficult to optimize as they often exhibit unstable training dynamics. 
Meanflow models exhibit excellent few-step sampling performance and tantalizing one-step sampling performance. 
Notably, MeanFlow models that achieve this have required extremely large training budgets.
We significantly decrease the amount of computation and data budget it takes to train Meanflow models by noting and exploiting a particular discretization of the Meanflow objective that yields a consistency property which we formulate into a ``Discrete Meanflow'' (DMF) Training Curriculum. 
Initialized with a pretrained Flow Model, DMF curriculum reaches one-step FID 3.36 on CIFAR-10 in only 2000 epochs. 
We anticipate that faster training curriculums of Meanflow models, specifically those fine-tuned from existing Flow Models, drives efficient training methods of future one-step examples. 
\end{abstract}

\section{Introduction}

Diffusion models and flow-based generative frameworks have fundamentally redefined the landscape of generative AI, offering a level of training stability and mode coverage that Generative Adversarial Networks (GAN) \cite{goodfellow2014generativeadversarialnetworks,karras2018progressivegrowinggansimproved} notably lacked \cite{ho2020denoisingdiffusionprobabilisticmodels,song2021scorebasedgenerativemodelingstochastic,lipman2023flowmatchinggenerativemodeling}.
By transforming noise priors into complex data distributions through a probability path, these models have demonstrated remarkable generalization across diverse modalities, including high-resolution images, video, and audio \cite{ho2022videodiffusionmodels,zhu2026audiogenerationscorebasedgenerative,podell2023sdxlimprovinglatentdiffusion}. 
However, unlike one-step GAN's, the inherent requirement of multi-step iterative sampling over the Probability Flow ODE path remains a significant bottleneck for real-time applications.
This limitation has sparked intense research interest in one/few-step variants, primarily through the lens of trajectory-aware distillation, reconciling high quality generation baselines with inference efficiency \cite{salimans2022progressivedistillationfastsampling,luhman2021knowledgedistillationiterativegenerative}.

The development of one/few-step generative models has increasingly shifted toward objectives derived from flow-matching trajectories. 
While initially shown to be an effective distillation technique to accelerate pre-trained teachers, researchers are increasingly interested in the potential of these objectives to function as self-contained, ``from scratch'' models. 
As a first, the self-supervised Consistency Training framework enabled Consistency Models to learn the solution of the underlying diffusion ODE \cite{song2023consistencymodels}.
This line of work demonstrated that one-step generation can be achieved by exploiting a fundamental consistency property between sample pairs along diffusion and flow trajectories \cite{kim2024consistencytrajectorymodelslearning,zheng2024trajectoryconsistencydistillationimproved,hu2025cmtmidtrainingefficientlearning}.

At the forefront of this shift is MeanFlow (MF) \cite{geng2025meanflowsonestepgenerative}, a framework that reformulated training around the average velocity over time intervals. 
Thanks to its ability to generate samples in one/few-steps, this approach has attracted follow-up works into more stabilized training dynamics, improvements in architecture, and distillation efficiency \cite{geng2025improvedmeanflowschallenges,lee2025decoupledmeanflowturningflow}. 
However, this theoretical elegance comes at a prohibitive cost: the continuous MF identity is notoriously expensive to train, often requiring heavy Jacobian-vector products (JVPs) that increases per-batch costs. 
Recent literature has sought to refine this; for instance, $\alpha$-Flow \cite{zhang2025alphaflowunderstandingimprovingmeanflow} explores the unification of moment matching with MF, while CMT \cite{hu2025meanflowtransformersrepresentationautoencoders} and iMT \cite{geng2025improvedmeanflowschallenges} explore the potential of integrating self-distillation targets directly into the MeanFlow objective to improve convergence. 
Much of the field’s recent progress has been driven by increasing model scale and computational resources, while comparatively less attention has been devoted to developing methods that make training more efficient and affordable \cite{geng2025improvedmeanflowschallenges}. 

In this work, we propose Discrete MeanFlow (DMF) training curriculum, a budget-friendly framework designed to bridge the gap between standard flow models and the MeanFlow identity for fast convergence. 
Our approach replaces expensive continuous identities with a staged curriculum that progressively introduces more challenging learning objectives. 
We demonstrate the efficacy of DMF on pixel-space CIFAR-10 \cite{krizhevsky2009learning}, achieving competitive FID \cite{heusel2018ganstrainedtimescaleupdate} scores at a fraction of the GPU-hour budget. 
On latent-space ImageNet $256\!\times\!256$ \cite{russakovsky2015imagenetlargescalevisual} we show that DMF scales effectively with increased training budget, exhibiting continuous performance improvements when initialized from a pretrained flow model. 
We further report findings on the stability ceiling observed in latent-space experiments, providing insights into how discretization granularity affects optimization robustness.

\section{Preliminary: The MeanFlow Identity}

MeanFlow (MF) proposes a framework for one-step generative modeling by shifting the perspective from instantaneous velocity fields, standard in flow matching, to average velocity formulation over time intervals. Consider a probability flow defined by the ordinary differential equation (ODE), \({\mathrm{d}\mathbf{z}_t}\!=\! \mathbf{v}_t(\mathbf{z}_t) \ \mathrm{d}t\), where \(\mathbf{z}_t \in \mathbb{R}^d \) represents the sample state at time \(t \in [0,1]\), \(\mathbf{z}_0 \sim p_\text{data}\) and \(\mathbf{z}_1 \sim \mathcal{N}(\mathbf{0}, \mathbf{I})\). Under a diffusion/flow-matching setting, the average velocity \(\mathbf{u}(\mathbf{z}_t, r, t)\) can be defined as the integrated displacement from time $r$ to $t$ divided by the interval $(t - r)$, with $0 \le r \le t \le 1$, i.e., $\mathbf{u}(\mathbf{z}_t, r, t) := \mathbf{z}_t - \mathbf{z}_r / (t-r)$. The \textit{MeanFlow Identity} yields,

\begin{equation}
\begin{aligned}
    \mathbf{u}(\mathbf{z}_t, r, t) &= \mathbf{v}_t(\mathbf{z}_t) + (r - t) \frac{\mathrm{d}}{\mathrm{d} t} \mathbf{u}(\mathbf{z}_t, r, t) \\
    &= \mathbf{v}_t(\mathbf{z}_t) + (r - t) \left(\frac{\partial \mathbf{u}(\mathbf{z}_t, r, t)}{\partial \mathbf{z}_t} \mathbf{v}_t(\mathbf{z}_t) +\frac{\partial \mathbf{u}(\mathbf{z}_t, r, t)}{\partial t} \right)
    \label{eq:meanflow_identity}
\end{aligned}
\end{equation}
For the complete derivation, we refer the reader to the original work \cite{geng2025meanflowsonestepgenerative}. In practice, MF models are trained to predict $\mathbf{u}_\theta(\mathbf{z}_t, r, t)$, where the training target is constructed by the conditional velocity field sampled at $\mathbf{z}_t$, and the Jacobian-vector product (JVP, \texttt{torch.func.jvp} in PyTorch) of the model with primals $(\mathbf{z}_t, r, t)$ and tangents $(\mathbf{v}_t, 0, 1)$. 

Alternatively, if we carry out the partial derivatives above by the definition of limits, we obtain their discrete forms, 

\begin{equation}
\label{eq:limit_def}
\begin{aligned}
\partial_{\mathbf{z}_t} \mathbf{u}(\mathbf{z}_t, r, t)
&= \left(
    \lim_{\|\boldsymbol{\delta}_i\| \to 0}
    \frac{
      \mathbf{u}(\mathbf{z}_t, r, t)_j
      - \mathbf{u}(\mathbf{z}_t - \boldsymbol{\delta}_i, r, t)_j
    }{\|\boldsymbol{\delta}_i\|}
   \right)_{i,j} := \mathcal{J}_{\mathbf{z}_t}, \\
\partial_t \mathbf{u}(\mathbf{z}_t, r, t) 
&= \lim_{\|\Delta\| \to 0}
   \frac{
     \mathbf{u}(\mathbf{z}_t, r, t)
     - \mathbf{u}(\mathbf{z}_t, r, t - \Delta)
   }{\|\Delta\|}.
\end{aligned}
\end{equation}
Plugging in the above limit definitions back into \eqref{eq:meanflow_identity}, we can derive the discretization,  
\begin{equation}
    \lim_{\Delta \to 0} \mathbf{u}(\mathbf{z}_t, r, t) = 
    \lim_{\Delta \to 0} \Biggl\{ 
    \frac{\mathbf{v}_t(\mathbf{z}_t) \cdot \Delta + \\
    \mathbf{u}(\mathbf{z}_t\!-\!\mathbf{v}_t \Delta , r, t\!-\!\Delta ) \cdot (t - r)}{(\Delta + t - r)} 
    \Biggr\},
\label{eq:discrete_meanflow}
\end{equation}
which we call \textit{Discrete MeanFlow} (DMF). A detailed derivation is provided in Appendix~\ref{a:proofs}. DMFs have been studied previously in attempt to unify the framework of Flow Models (FM) to MeanFlows \cite{zhang2025alphaflowunderstandingimprovingmeanflow}, as well as approximating the convergence of MFs without computing the JVP \cite{hu2025meanflowtransformersrepresentationautoencoders}. If r is fixed, the form in \eqref{eq:discrete_meanflow} reveals a consistency property that aligns the average velocity from different samples along the trajectory that is corrected by the instant velocity change. In practice, DMF model predicts $\mathbf{u}_\theta(\mathbf{z}_t, r, t)$, and the target is simply the interpolation between $\mathbf{v}_t$ and $\text{sg} (\mathbf{u}_\theta(\mathbf{z}_t\!-\!\mathbf{v}_t \Delta, r, t\!-\!\Delta))$, with sg($\cdot$) denoting the stop gradient. In our work, we study the benefits and stability of decreasing the $\Delta$ term as a step function. This approach is motivated by the success of training curriculums in Consistency Training Models \cite{song2023improvedtechniquestrainingconsistency,geng2024consistencymodelseasy,dao2025improvedtrainingtechniquelatent}. We provide our detailed methodology in the following section. 

\section{Method: Training Curriculum}

Notice that $\mathbf{v}_t$ in both DMF and MF attributes to the signal of convergence. Without it, the model would collapse to a simple solution that produces arbitrary constant. Drawing inspiration from training curriculums in Consistency Models, we propose Discrete MeanFlow (DMF) Training Curriculum that aims to improve convergence of the consistency property by adaptively \textit{turning down the knob}, $\Delta$, in \eqref{eq:discrete_meanflow}. Following ECT \cite{geng2024consistencymodelseasy}, our training curriculum equally divides the target training budget into different stages, each with different training targets denoted as $\mathbf{u}^i_\text{target}, i \in \{ 0,...,K\!-\!1 \}$, where $i$ is the stage of training, and $K$ is the total number of stages.

We start with a large $\Delta$ at the beginning of the curriculum. The step size $\Delta$ in DMF can be as large as $(t\!-\!r)$. In this case, the target for $\mathbf{u}_\theta({\mathbf{z}_t,r,t})$ becomes $\frac{1}{2}(\mathbf{v}_t(\mathbf{z}_t)+\frac{1}{2}\text{sg}(\mathbf{u}_\theta(\mathbf{z}_r,r,r))$, where $\mathbf{z}_r$ is the cleaner sample that lies on the linear trajectory induced by the conditional velocity field $\mathbf{v}_t = \mathbf{\epsilon} - \mathbf{z}_0, \ \mathbf{z}_0\!\sim\!p_\text{data} $ and $\mathbf{\epsilon}\!\sim\!\mathcal{N}(\mathbf{0}, \mathbf{I})$. Observe that the ground truth for $\mathbf{u}_\theta(\mathbf{z}_r,r,r)$ is the instantaneous velocity at $r$ \cite{geng2025meanflowsonestepgenerative}, which coincidentally, is also trained using the conditional velocity. Therefore, our first stage of the curriculum collapses to the flow matching objective, where the target is simply $\mathbf{u}_\text{target}^0 := \mathbf{v}_t$.

For the following $i$-th intermediate stages, $i \in \{1,...,K\!-\!2\} $, we adaptively decrease the step size $\Delta$ by defining it as a function of stage, denoted as $\Delta_i$. With a chosen shrinking factor $q$, a straightforward design choice would suggest $\Delta_i=(t-r)/q^i$. However, we found out that this led to suboptimal FID performance from our CIFAR-10 experiments. We discovered that a training curriculum based on a noise schedule mapped to the \textit{Variance-Exploding} (VE) diffusion framework \cite{song2021scorebasedgenerativemodelingstochastic} proved particularly effective. Specifically, let $\Phi(t) = t/(1-t)$ be the transformation that maps time $t$ to the VE scheme, then

\begin{equation} 
    t' = \Phi^{-1}\left( \Phi(t) - \frac{\Phi(t) - \Phi(r)}{q^i} \right), 
    \quad \Delta^{\dagger}_i = t - t'.
\end{equation}

As a result, the target for these intermediate stages follows as substituting the $\Delta$ with $\Delta^{\dagger}_i$ in the DMF \eqref{eq:discrete_meanflow}. For the last stage (stage $K\!-\!1$), we fallback to train the model with the MF objective, with the assumption this smoothly transitions to the \textit{hardest} objective. In summary, our training curriculum defines a sequence of objectives for different stages as,
\begin{equation}
    \mathbf{u}_\text{target}^i := 
    \begin{cases}
        \mathbf{v}_t, & \text{if } i = 0, \\[1em]
        
        \dfrac{\mathbf{v}_t \cdot \Delta^{\dagger}_i + \text{sg} \left(\mathbf{u}_\theta(\mathbf{z}_t\!-\!\mathbf{v}_t\!\cdot\!\Delta^{\dagger}_i, r, t\!-\!\Delta^{\dagger}_i)\right) \cdot (t - r)}{\Delta^{\dagger}_i + t - r}, & \text{if } 1\le i \le K\!-\!2, \\[1em]
        
        \mathbf{v}_t + (r-t) \cdot \text{sg}\left(\frac{\mathrm{d}}{\mathrm{d} t} \mathbf{u}_\theta(\mathbf{z}_t, r, t)\right), & \text{if } i = K\!-\!1.
    \end{cases}
    \label{eq:target_objectives}
\end{equation}
Note that we follow MF models to compute the JVP with PyTorch's forward auto-differentiation operation in the last stage. A detailed full procedure of the DMF curriculum is provided in the Appendix~\ref{a:alg_overview}. 

\begin{figure}[t]
    \centering
    \includegraphics[width=0.8\linewidth]{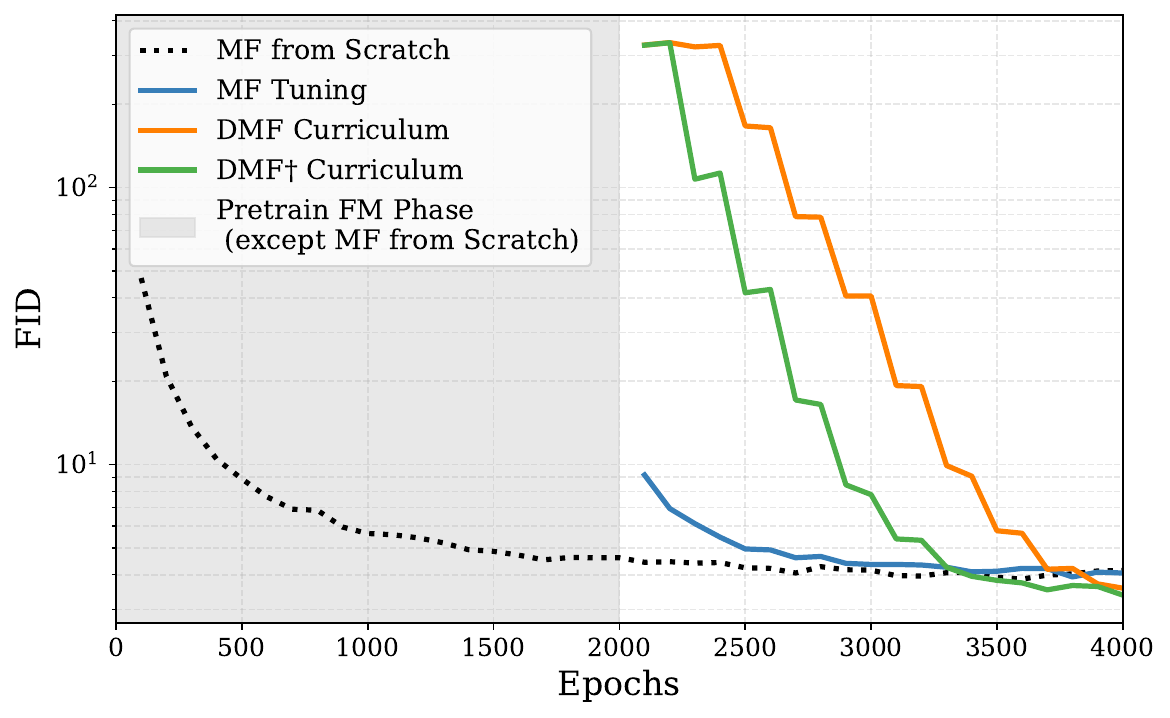}
    \caption{Training convergence on unconditional CIFAR-10. DMF curriculums achieve better 1-step FID compared to the MF baseline with equal training data budget, despite starting from a pretrained flow model at 2000 epochs.}
    \label{fig:cifar10_convergence}
\end{figure}

\section{Experiments}

DMF training curriculum shares the same target with flow-matching at the first stage of training. This suggests that initializing our model with a pretrained Flow Model (FM) is a better candidate than random initialization. Our baseline for the CIFAR-10 experiments implements the configurations from the official MF paper. We strictly follow Appendix A in \cite{geng2025meanflowsonestepgenerative}, with the only difference being EMA ratio that is scaled down w.r.t. the limited training budget. To demonstrate the efficacy of our curriculum under a cold-start scenario, we impose a fixed budget of 4000 epochs for models trained from random initialization and report the best FID achieved. For fair comparison, experiments utilizing MF fine-tuning or the DMF curriculum initialized from an FM are restricted to a 2000-epoch budget, as the base FM itself has been pretrained for 2000 epochs. Our pretrained FM follows the standard configurations from the official flow matching repository\footnote{\texttt{github.com/facebookresearch/flow\_matching}}., which achieves an FID of 3.09 with 50-steps sampling from our reimplementation. 

Following the latent diffusion paradigm, ImageNet 256$\times$256 samples are compressed via the SD-VAE \cite{rombach2022highresolutionimagesynthesislatent} into a 4×32×32 latent space prior to training.  We initialize the model from the weights of a pretrained SiT-XL/2 \cite{ma2024sitexploringflowdiffusionbased}. Due to computational constraints, we evaluate the DMF curriculum on image generation without Classifier-Free Guidance (CFG) \cite{ho2022classifierfreediffusionguidance}. We omit CFG for two primary reasons: first, it significantly increases training overhead by requiring two additional model passes to compute the interpolated target (Tab.~\ref{tab:cost_batch}); second, determining the optimal guidance scale w is computationally expensive, requiring extensive hyperparameter sweeps. Consequently, we compare our results against the SiT-XL/2 baseline sampled without CFG to ensure a fair, though unorthodox, comparison within a fixed training budget.

\subsection{Unconditional CIFAR-10}

\textbf{Training Configuration.} The baselines include MeanFlow (MF) trained from scratch and MF fine-tuned from a pretrained Flow Model (FM). We set the number of stages for the DMF curriculum to $K=10$, with a decay factor $q = 2$. We distinguish DMF and DMF$^{\dagger}$ by their curriculum schedules $\{\Delta_i\}$ and $\{\Delta_i^{\dagger}\}$, respectively, as defined in Section 3. Besides, both DMF's are trained with approximately 100\% MF objective. Only regions where timestep difference is infinitesimally small, \(t\!-\!r\!<\!\epsilon_t\), $\epsilon_t\!=\!10^{-6}$, the target is set to the velocity field $\mathbf{v}_t$ for numerical stability. All MF and DMF models are trained using the Adam optimizer with a learning rate of $6\times10^{-4}$, batch size 1024, adaptive loss \cite{geng2024consistencymodelseasy}, and are evaluated using the same EMA=0.999. A detailed configuration is included in Appendix~\ref{a:configs}. 

\textbf{CIFAR-10 Convergence Analysis.} Consistent with observations from curriculum training in Consistency Models \cite{geng2024consistencymodelseasy}, MF training and fine-tuning exhibit faster initial convergence, but their improvements diminish at later stages. As shown in Fig.~\ref{fig:cifar10_convergence}, MF achieves a final FID of 3.85 when trained from scratch and 3.93 when fine-tuned from an FM initialization. In contrast, DMF curriculum training progresses more slowly and improves in a stage-wise manner. The discontinuity in FID improvement reflects convergence within each curriculum stage. Despite this slower early progress, DMF achieves superior final performance, with DMF$^{\dagger}$ using the VE-transformed scheduler reaching a comparable FID of 3.36. Notably, DMF curriculum training attains competitive or better FID compared to prior methods trained with substantially larger computational budgets (Tab.~\ref{tab:unified_results}).

\begin{table}[t]
    \centering
    \small
    \caption{\textbf{CIFAR-10 comprehensive 1-step FID comparison}. Methods are categorized by initialization strategy. Budgets represent total training epochs, for models initialized with pretrained models, the budget are formatted as ``Cost of Pretraining'' + ``Cost of Tuning''. DMF curriculum training attains competitive or better FID compared to prior methods.}
    \label{tab:unified_results}
    \begin{tabular}{@{}llcc@{}}
        \toprule
        \textbf{Method} & \textbf{Initialization} & \textbf{Budget (Epochs)} & \textbf{FID ($\downarrow$)} \\ 
        \midrule
        \rowcolor{gray!10} \multicolumn{4}{c}{\textit{From Scratch}} \\
        iCT & Random & 8k & \textbf{2.83} \\
        MF (ours imple.) & Random & 4k & 3.85 \\
        MF \cite{geng2025meanflowsonestepgenerative} & Random & 16k & \underline{2.90} \\
        \midrule
        \rowcolor{gray!10} \multicolumn{4}{c}{\textit{With DM / FM Initialization}} \\
        sCT \cite{lu2025simplifyingstabilizingscalingcontinuoustime} & Pretrained DM & 4k + 4k & \textbf{2.85} \\
        ECT \cite{geng2024consistencymodelseasy} & Pretrained DM & 4k + 1k & 3.60 \\
        MF (ours imple.) & Pretrained FM & 2k + 2k & 3.93 \\
        DMF Curriculum & Pretrained FM & 2k + 2k & 3.58 \\
        DMF$^\dagger$ Curriculum & Pretrained FM & 2k + 2k & \underline{3.36} \\
        \bottomrule
        \addlinespace
        \multicolumn{4}{l}{\scriptsize $^\dagger$ Denotes DMF Curriculum using the VE-transformed scheduler.}
    \end{tabular}
\end{table}


\begin{table}[t]
\centering
\small
\begin{minipage}[t]{0.48\linewidth}
\centering
\caption{\textbf{Per-batch training cost}. Batch size 1024, 4 H100's. The DMF loss is approximately \textcolor{blue}{\textbf{1.2$\times$}}$\sim$\textcolor{blue}{\textbf{1.8$\times$}} faster than MF as it does not require heavy JVP. The cost of MF is computed without doing classifier-free guidance, so the scale ratio has further been reduced from MF in practice, i.e., excluding 2 extra forward passes.}
\label{tab:cost_batch}
\vspace{5pt} 
\begin{tabular}{llc}
\toprule
Dataset & Method & Sec.$/$Batch \\
\midrule
CIFAR-10 & MF & 0.38 \\
& \textbf{DMF} & \textbf{0.32} \\
\midrule
ImageNet $256\!\times\!256$ & MF & {3.08}  \\
& \textbf{DMF} & \textbf{1.71} \\
\bottomrule
\end{tabular}
\end{minipage}
\hfill
\begin{minipage}[t]{0.48\linewidth}
\centering
\caption{\textbf{CIFAR-10 end-to-end training cost for same data budget and final FID.} DMF Curriculum consists of 2000 epochs of Flow Model training followed by 2000 epochs of DMF tuning, and is compared against MeanFlow trained from scratch for 4000 epochs. In terms of GPU hours measured in H100's under the same data budget, DMF Curriculum \textcolor{blue}{\textbf{1.3$\times$}} faster.}
\label{tab:mf_cost_total}
\vspace{5pt} 
\begin{tabular}{lcc}
\toprule
Method & GPU Hours & FID \\
\midrule
MF & 85.33 & 3.85 \\
\textbf{DMF Curriculum} & \textbf{66.6} & \textbf{3.36} \\
\bottomrule
\end{tabular}
\end{minipage}
\end{table}

\subsection{ImageNet 256x256, SD-VAE latents}

\textbf{Training Configuration.} To evaluate the scalability of our approach, we apply the DMF curriculum to a SiT-XL/2 baseline pretrained for 1400 epochs on the SD-VAE latents, which achieves an FID of 11.52 with 50-step sampling without Classifier-Free Guidance (CFG). Latent spaces encoded via SD-VAE are known to be susceptible to extreme outliers at Consistency Training \cite{dao2025improvedtrainingtechniquelatent,hu2025meanflowtransformersrepresentationautoencoders}. To mitigate this, we employ a robust Cauchy loss with a high robust value of $c=0.3$. Plus, instead of using the conditional velocity as $\mathbf{v}_t$, we compute the velocity field using softmax as a kernel \cite{xu2023stabletargetfieldreduced} by sampling an additional sub-batch of 127 data from the dataset for each sample, reducing the variance at training. A detail note on this will be provided in Appendix~\ref{a:alg_overview}. We utilize a 6-stage $\Delta^{\dagger}_i$ curriculum ($K=6$) with a factor decay of $q\!=\!4$. 

\textbf{Direct Fine-tuning from Pure MeanFlow.} As shown in Tab.~\ref{tab:imagenet_budget}, DMF$^\dagger$ yields rapid convergence in the low-epoch regime, reaching FID of 21.18 with a training budget of 6 epochs, and to 14.53 with increased training budget of 48 epochs. A significant departure from standard practice is our use of a ``pure'' MeanFlow objective. While existing methods typically rely on a hybrid training mixture of flow-matching and MF (MF) objectives in ratios ranging from 1:1 to 3:1 \cite{geng2025improvedmeanflowschallenges,geng2025meanflowsonestepgenerative, zhang2025alphaflowunderstandingimprovingmeanflow,hu2025meanflowtransformersrepresentationautoencoders}, we perform direct fine-tuning using a nearly 100\% MF objective (Tab.~\ref{tab:pure_vs_hybrid}). Similar to the CIFAR-10 experiments, the only regions where we do flow-matching is where the timestep difference is infinitesimal, $t\!-\!r\!<\epsilon_t$. 

\textbf{Stability Analysis and Optimization Limits.} Despite the aforementioned robustness measures, we observed an empirical stability ceiling when scaling the DMF$^{\dagger}$ training budget to 96 epochs. Specifically, optimization tends to diverge during the fifth curriculum stage, where the discretization ratio reaches approximately $\Delta^\dagger_4 \approx 0.0039\cdot(t-r)$. We hypothesize that for latent-space datasets, excessively fine discretization introduces a critical trade-off between approximation accuracy and training stability. This suggests that intermediate stages with high discretizations may act as a source of variance that destabilizes the objective. Potential remedies for this instability include an early transition to the MF regime during intermediate stages, or a more granular analysis of model architecture and normalization layers to improve robustness under small step-size regimes.

\begin{table}[t]
    \centering
    \caption{\textbf{ImageNet $256\!\times\!256$ curriculum training budget w.r.t FID.} We report the FID of our 1-step DMF curriculum relative to the 1400-epoch pretrained SiT-XL/2 baseline. The FID is computed on samples generated w/o CFG sampling or tuning.}
    \label{tab:imagenet_budget}
    \begin{tabular}{@{}llccc@{}}
        \toprule
        \textbf{Method} & \textbf{Training Epochs} & \textbf{Rel. Budget} & \textbf{Steps} & \textbf{FID $\downarrow$} \\ 
        \midrule
        SiT-XL/2 (Baseline) & 1400 (Pretrain) & 100.0\% & 50 & 11.52 \\
        \midrule
        DMF$^\dagger$ & 1400 + 6 & +0.42\% & 1 & 21.18 \\
        DMF$^\dagger$ & 1400 + 12 & +0.85\% & 1 & 18.03 \\
        DMF$^\dagger$ & 1400 + 24 & +1.71\% & 1 & 16.95 \\
        DMF$^\dagger$ & 1400 + 48 & +3.42\% & 1 & 14.53 \\
        DMF$^\dagger$ & 1400 + 96 & +3.42\% & 1 & 294.13 \\
        \bottomrule
    \end{tabular}
\end{table}

\begin{table}[t]
    \centering
    \caption{\textbf{Comparison of Training Paradigms on ImageNet-256.} Unlike majority MF that are trained on hybrid objectives or distillation teachers, our DMF curriculum enables ``pure'' MF fine-tuning.}
    \label{tab:pure_vs_hybrid}
    \begin{tabular}{@{}lccc@{}}
        \toprule
        \textbf{Strategy} & \textbf{Objective Ratio (FM:MF)} & \textbf{Training Type} & CFG \\ 
        \midrule
        Standard MF \cite{geng2024consistencymodelseasy} & 1:1 to 3:1 & Direct Training & Yes \\
        RAE-MF \cite{hu2025meanflowtransformersrepresentationautoencoders} & 3:1 & Mid-training Teacher & None \\
        $\alpha$-Flow \cite{zhang2025alphaflowunderstandingimprovingmeanflow} & 1:1 & Direct Curriculum & Yes \\ 
        \midrule
        DMF Curriculum & 0:1 (Pure MF) & Direct Curriculum & None \\ 
        \bottomrule
    \end{tabular}
\end{table}

\section{Conclusion and Limitations}

In this paper, we show that Discrete MeanFlow (DMF) training curriculum enables high quality one step generation by replacing continuous MeanFlow identities with a staged curriculum of discrete approximations. Experiments on CIFAR-10 and ImageNet $256\!\times\!256$ show that DMF achieves comparable FID with chosen baselines under fixed data budgets, while achieving accelerated convergence as up to 1.8$\times$ per batch speedup by avoiding the expensive JVP computations. We further analyze training stability and identify an empirical discretization ceiling in the latent space. When the curriculum becomes too fine, optimization can diverge, revealing a trade off between progressive discretization and training robustness. Future work should focus on enhancing the scalability and robustness of the training curriculum framework through targeted architectural and procedural improvements. Specifically, evaluating it on Representation-learning Autoencoder (RAE) latents \cite{zheng2025diffusiontransformersrepresentationautoencoders} could determine curriculum's effectiveness and stability on more structured manifolds. Furthermore, introducing a lightweight secondary guidance tuning stage that isolates classifier free guidance allows the primary training phase to remain budget friendly. Finally, investigating architecture-specific robustness, such as the role of normalization layers and weight initialization, remains critical for mitigating the optimization divergence observed in the final stages of latent-space training.

\subsubsection*{Acknowledgments}
We thank Yingchen He for running the experiments, Matthew Niedoba for suggestions on stabilizing the velocity field, and Saeid Naderiparizi for helpful discussions and feedback on the paper.

\bibliography{iclr2026_delta}
\bibliographystyle{iclr2026_delta}

\newpage

\appendix
\section{Appendix}

\subsection{Proofs}
\label{a:proofs}

\begin{proof}

To keep the proof of~\eqref{eq:discrete_meanflow} elegant, we simplify the notation by setting $r=0$ starting from the MeanFlow Identity (~\eqref{eq:meanflow_identity}), and omit it during our derivations. We will add $r$ back to match the generalized discretized form. Starting from the MF Identity, we have:
\begin{equation}
\begin{aligned}
    \mathbf{u}(\mathbf{z}_t, t) &= \mathbf{v}_t(\mathbf{z}_t) - t \frac{\mathrm{d}}{\mathrm{d} t} \mathbf{u}(\mathbf{z}_t, t) \\
    \\
    &= \mathbf{v}_t(\mathbf{z}_t) - t \left(\frac{\partial \mathbf{u}(\mathbf{z}_t, t)}{\partial \mathbf{z}_t} \mathbf{v}_t(\mathbf{z}_t) +\frac{\partial \mathbf{u}(\mathbf{z}_t, t)}{\partial t} \right) \\
    \\
    &= \mathbf{v}_t(\mathbf{z}_t) - t \  \left( \ \mathcal{J}_{\mathbf{z}_t} \mathbf{v}_t (\mathbf{z}_t,t) + \partial_t \mathbf{u}(\mathbf{z}_t, t) \ \right) \\
    \\
    &= \mathbf{v}_t(\mathbf{z}_t) - t \left( \lim_{\Delta \to 0} \frac{\mathbf{u}(\mathbf{z}_t, t) - \mathbf{u}(\mathbf{z}_t\!-\!\mathbf{v}_t\Delta, t)}{ \Delta} + \lim_{\Delta \to 0} \frac{\mathbf{u}(\mathbf{z}_t, t) - \mathbf{u}(\mathbf{z}_t, t\!-\!\Delta)}{\Delta} \right) \\
    \\
    &\qquad (\text{merge partial limits into the total derivative along the trajectory } \mathbf{z}_t(t)) \\
    \\
    &= \mathbf{v}_t(\mathbf{z}_t) - t \left( \lim_{\Delta \to 0}  \frac{\mathbf{u}(\mathbf{z}_t, t) - \mathbf{u}(\mathbf{z}_t\!-\!\mathbf{v}_t\Delta, t\!-\!\Delta)}{\Delta} \right)
\end{aligned}
\end{equation}
Multiply by $\Delta$ to clear the denominator, both sides.
\begin{align*}
    \lim_{\Delta \to 0} \left[ \mathbf{u}(\mathbf{z}_t,t) \cdot \Delta \right] &= \lim_{\Delta \to 0} \left[ \mathbf{v}_t(\mathbf{z}_t) \cdot \Delta - t \cdot \left( \mathbf{u}(\mathbf{z}_t, t) - \mathbf{u}(\mathbf{z}_t\!-\!\mathbf{v}_t\Delta , t\!-\!\Delta) \right) \right]
\end{align*}
Move $\mathbf{u}(\mathbf{z}_t, t)$ to the L.H.S.
\begin{align*}
    \lim_{\Delta \to 0} \left[ \mathbf{u}(\mathbf{z}_t,t) \cdot (\Delta+t) \right] = \lim_{\Delta \to 0} \left[  \mathbf{v}_t(\mathbf{z}_t) \cdot \Delta + t \cdot \mathbf{u}(\mathbf{z}_t\!-\!\mathbf{v}_t\Delta , t\!-\!\Delta) \right]
\end{align*}
Divide both sides $(\Delta+t)$.
\begin{align*}
    \mathbf{u}(\mathbf{z}_t,t) &= \lim_{\Delta \to 0} \frac{\mathbf{v}_t(\mathbf{z}_t) \cdot \Delta + t \cdot \mathbf{u}(\mathbf{z}_t\!-\!\mathbf{v}_t\Delta , t\!-\!\Delta)}{\Delta+t}  
\end{align*}
Bring back $r$, we get the final discretized form.
\begin{align*}
    \mathbf{u}(\mathbf{z}_t,r,t) &= \lim_{\Delta \to 0} \frac{\mathbf{v}_t(\mathbf{z}_t) \cdot \Delta + (t -r)\cdot \mathbf{u}(\mathbf{z}_t\!-\!\mathbf{v}_t\Delta , r, t\!-\!\Delta)}{\Delta+t-r}  
\end{align*}
\end{proof}

\subsection{Algorithm Overview}
\label{a:alg_overview}

\begin{algorithm}[H]
\caption{Discrete MeanFlow (DMF) Curriculum Training }
\label{alg:dmf}
\begin{algorithmic}[1]
   \STATE {\bfseries Input:} Pretrained flow model $\mathbf{v}_{\phi}$, dataset $\mathcal{D}$, total stages $K$, decay factor $q$, robust value $c$, sub-batch size $B_\text{sub}$, $\texttt{LogitNormal}$ hyperparams $P_\text{mean}$, $P_\text{std}$, training epochs per stage $N_\text{epochs}$.
   \STATE {\bfseries Initialize:} Model $\mathbf{u}_{\theta} \leftarrow \mathbf{v}_{\phi}$.
   \FOR{$i = 0$ {\bfseries to} $K-1$}

   \FOR{$n=0$ {\bfseries to} $N_\text{epochs}-1$}
    \STATE Sample $\mathbf{z}_0 \sim \mathcal{D}, \mathbf{\epsilon} \sim \mathcal{N}(\mathbf{0}, \mathbf{I}), t,r \sim \texttt{LogitNormal}(P_\text{mean},P_\text{std})$.
    \STATE Compute $\mathbf{z}_t \leftarrow (1-t) \mathbf{z}_0 + t \mathbf{\epsilon}$, $\mathbf{z}_r \leftarrow (1-r) \mathbf{z}_0 + r \mathbf{\epsilon}$. 
    \STATE 
    \STATE Compute velocity field.
    \IF{$\mathcal{D}$ is ImageNet $256\!\times\!256$}
            \STATE Sample subset $\mathcal{X}_{sub} \leftarrow \{\mathbf{x}_0^{(k)}\}_{k=1}^{B_\text{sub}}$ from the same class as $\mathbf{x}_0$.
            \STATE Include current sample in the reference set $\mathcal{X}_{sub} \leftarrow \mathcal{X}_{sub}' \cup \{\mathbf{x}_0\}$. 
        \STATE Compute normalized weights via softmax over $B_\text{sub}\!+\!1$ samples.
        \FOR{$k = 0$ {\bfseries to} $B_\text{sub}$}
            \STATE $w_k \leftarrow \frac{\exp\left( -\|\mathbf{z}_t - (1-t)\mathbf{x}_0^{(k)}\|^2 / (2t^2) \right)}{\sum_{j=0}^{B_\text{sub}} \exp\left( -\|\mathbf{z}_t - (1-t)\mathbf{x}_0^{(j)}\|^2 / (2t^2) \right)}$.
        \ENDFOR
        \STATE Compute stable reference: $\bar{\mathbf{x}}_0 \leftarrow \sum_{k=0}^{B_\text{sub}} w_k \mathbf{x}_0^{(k)}$.
        \STATE $\mathbf{v}_t \leftarrow \frac{\mathbf{z}_t - \bar{\mathbf{x}}_0}{t}$ \COMMENT{Stable target field \cite{xu2023stabletargetfieldreduced}}.
    \ELSE
        \STATE $\mathbf{v}_t \leftarrow \mathbf{\epsilon} - \mathbf{z}_0 $.
    \ENDIF
    \STATE 
    \STATE Compute $\mathbf{u}_\text{target}$.
    \IF{$i = 0$}
        \STATE $\mathbf{u}_\text{target} \leftarrow \mathbf{v}_t$.
    \ELSIF{$i = K-1$}
        \STATE Compute $ \_, \ \text{dudt} \leftarrow \texttt{jvp}( (\mathbf{z}_t,r,t), (\mathbf{v}_t,0,1))$.
        \STATE $\mathbf{u}_\text{target} \leftarrow \text{sg} \left\{ \mathbf{v}_t - (t-r) \cdot \text{dudt} \right\}$.
    \ELSE
        \STATE Compute $\Phi(t) \leftarrow  t/(1-t), \Phi(r) \leftarrow r/(1-r)$. 
        \STATE Compute $\Delta^{\dagger}_i  \leftarrow t - \Phi^{-1}(\Phi(t) - 1/q^i \cdot (\Phi(t)-\Phi(r))) $.
        \STATE $\mathbf{u}_\text{target} \leftarrow \text{sg}\left\{ \frac{1}{(\Delta_i^{\dagger} + t - r)} \cdot \left[ \mathbf{v}_t \Delta_i^{\dagger} + \mathbf{u}_{\theta}(\mathbf{z}_t\!-\!\mathbf{v}_t\Delta_i^{\dagger}, t\!-\!\Delta_i^{\dagger})(t-r) \right] \right\}$
    \ENDIF
    \STATE 
    \IF{$\mathcal{D}$ is ImageNet $256\!\times\!256$}
        \STATE $\text{Loss} \leftarrow \text{Cauchy}(\cdot,c)$.
    \ELSE
        \STATE $\text{Loss} \leftarrow \text{Adaptive}(\cdot,c) $ \cite{geng2025meanflowsonestepgenerative}.
    \ENDIF
    \STATE 
    \STATE $\mathcal{L}(\theta) \leftarrow \text{Loss}(\mathbf{u}_{\theta}(\mathbf{z}_t, r,t), \mathbf{u}_\text{target})$
    \STATE Update $\theta$ via gradient descent
    \ENDFOR
\ENDFOR
   \STATE {\bfseries Return:} Optimized 1-step model $\mathbf{u}_{\theta}$
\end{algorithmic}
\end{algorithm}

\subsection{Configurations}
\label{a:configs}

\begin{table}[H]
\centering
\caption{CIFAR-10, hyperparameter configurations across different experimental setups.}
\label{tab:hyperparams}
\small 
\begin{tabular}{@{}lcccc@{}}
\toprule
\textbf{Hyperparameter} & \textbf{FM Pretrain} & \textbf{MF (Scratch)} & \textbf{MF (Fine-tune)} & \textbf{DMF$^\dagger$ curriculum} \\ 
\midrule
Batch Size             &     256             & 1024                  & 1024                    & 1024                \\
Training Epochs  & 2000                  & 4000                   & 2000                      & 2000           \\
Optimizer              &   Adam            & Adam                  & Adam                    & Adam                \\
Learning Rate          &  $2\times10^{-4}$     & $6\times10^{-4}$      & $6\times10^{-4}$        & $6\times10^{-4}$    \\
EMA Rate               & 0.999                & 0.999                 & 0.999                   & 0.999               \\
Network Dropout       &      0.2          & 0.2            & 0.2                 & 0.35              \\
Stages ($K$)           & N/A                  & N/A                   & N/A                     & 10                  \\
Decay Factor ($q$)     & N/A                  & N/A                   & N/A                     & 2                   \\
$\epsilon_t$          & N/A        & N/A       & N/A       & $10^{-6}$   \\ 
Loss Function          & Mean Squared Error   & Adaptive $L_p$        & Adaptive $L_p$          & Adaptive $L_p$     \\ 
Adaptive norm\_p       & N/A                  & 0.75       & 0.75         & 0.75     \\ 
Adaptive c             & N/A                  & 0.001       & 0.001         & 0.001     \\ 
Logitnormal $P_\text{mean}$            & -1.2                 & -2.0       & -2.0       & -2.0    \\ 
Logitnormal $P_\text{std}$            & 1.2                & 2.0      & 2.0        & 2.0   \\ 
Probability $t$ equal $r$           & N/A        & 0.25      & 0.25       & 0   \\ 
\bottomrule
\end{tabular}
\end{table}
\begin{table}[H]
\centering
\caption{ImageNet $256\!\times\!256$ hyperparameter configurations across different experimental setups.}
\label{tab:hyperparams_imagenet}
\resizebox{\textwidth}{!}{ 
\begin{tabular}{@{}lccccc@{}}
\toprule
\textbf{Hyperparameter} & \textbf{DMF$^\dagger$ 6ep} & \textbf{DMF$^\dagger$ 12ep} & \textbf{DMF$^\dagger$ 24ep} & \textbf{DMF$^\dagger$ 48ep} & \textbf{DMF$^\dagger$ 96ep} \\ 
\midrule
Batch Size             & 1024                  & 1024                 & 1024                 & 1024                 & 1024                 \\
Training Epochs  & 6                    & 12                   & 24                   & 48                   & 96                   \\
Optimizer              & AdamW                 & AdamW                & AdamW                 & AdamW                 & AdamW                 \\
Learning Rate          & $1\times10^{-4}$     & $1\times10^{-4}$     & $1\times10^{-4}$     & $1\times10^{-4}$     & $1\times10^{-4}$     \\
EMA Rate               & 0.995                & 0.999                & 0.999                & 0.9995                & 0.9995                \\
Network Dropout        & 0.0                  & 0.0                  & 0.0                  & 0.0                 & 0.0                 \\
Stages ($K$)           & 6                   & 6                   & 6                   & 6                   & 6                   \\
Decay Factor ($q$)     & 4                    & 4                    & 4                    & 4                    & 4                    \\
$\epsilon_t$          & $10^{-3}$       & $10^{-3}$      & $10^{-3}$      & $10^{-3}$ & $10^{-3}$   \\ 
Loss Function          & Cauchy                  & Cauchy      & Cauchy      & Cauchy       & Cauchy       \\ 
Robust $c$           & 0.3                  & 0.3               & 0.3              & 0.3                & 0.3                \\ 
Logitnormal $P_\text{mean}$ & -0.4            & -0.4                  & -0.4                   & -0.4                    & -0.4                  \\ 
Logitnormal $P_\text{std}$  & 1.0             & 1.0                  & 1.0                 & 1.0                     & 1.0                    \\ 
Prob. $t=r$            & 0.0                & 0.0              & 0.0                 & 0.0                    & 0.0                    \\ 
\bottomrule
\end{tabular}
}
\end{table}
\subsection{Uncurated Samples CIFAR-10}

\begin{figure}[H]
    \centering
    \includegraphics[width=0.9\textwidth]{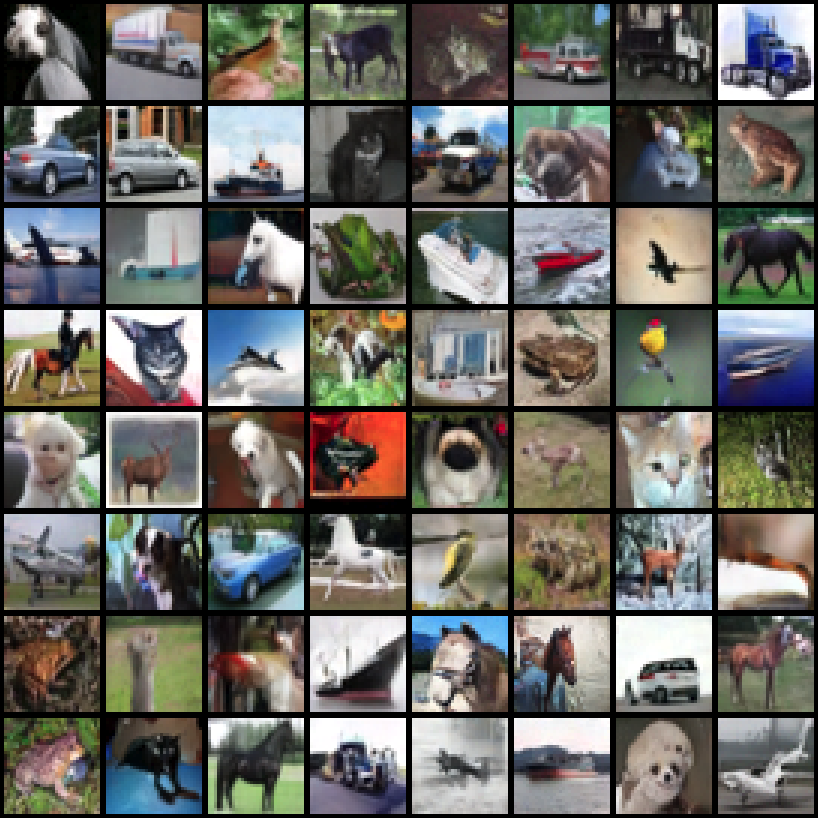}
    \caption{DMF$^{\dagger}$ on CIFAR-10, FID=3.36.}
    \label{fig:uncurated_cifar}
\end{figure}

\begin{figure}[H]
    \centering
    \includegraphics[width=0.9\textwidth]{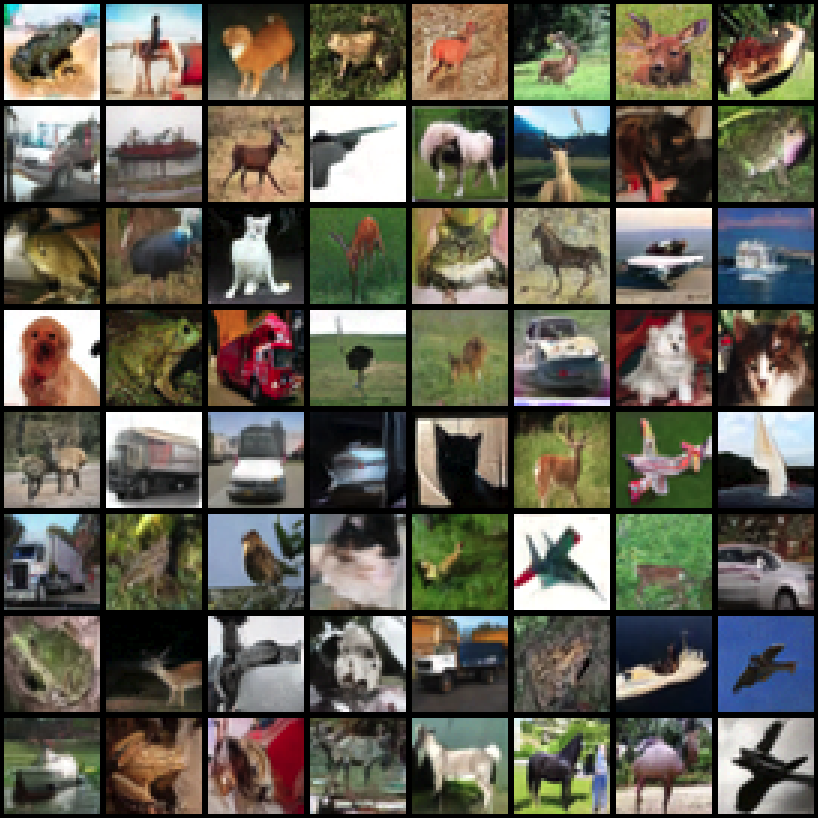}
    \caption{DMF$^{\dagger}$ on CIFAR-10, FID=3.36.}
    \label{fig:uncurated_cifar}
\end{figure}

\begin{figure}[H]
    \centering
    \includegraphics[width=0.9\textwidth]{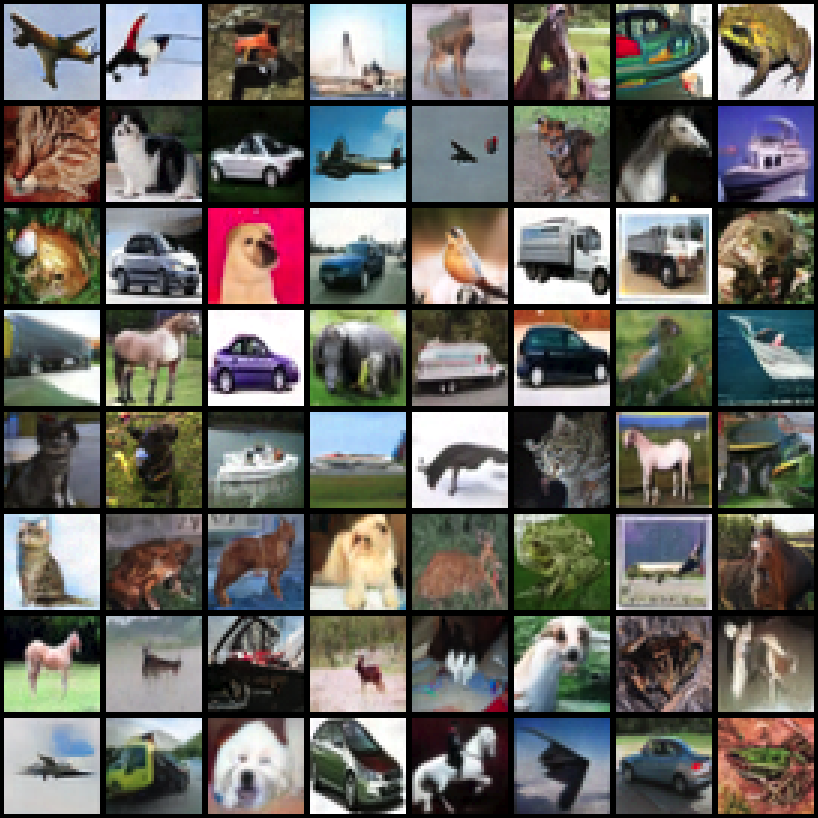}
    \caption{DMF$^{\dagger}$ on CIFAR-10, FID=3.36.}
    \label{fig:uncurated_cifar}
\end{figure}

\end{document}